%% file: main.tex
\documentclass[letterpaper, 10 pt, conference]{ieeeconf}  % Comment this line out
\IEEEoverridecommandlockouts                              % This command is only
\usepackage[autonum,colorhypersetup]{shortex} 

\usepackage{stackengine}

 \definecolor{mpcritic}{RGB}{10, 92, 174}
 \definecolor{warmstart}{RGB}{236, 100, 75}
 
\usepackage{graphicx}
\graphicspath{{figures/}}

\usepackage{acro}
\acsetup{single={1},	
	uppercase/list}

\input{./acronyms} % (this is just to make my life easier when transferring content)

\usepackage{algorithm}
\usepackage[noend]{algorithmic}
\usepackage[scr=boondox]{mathalfa}

\usepackage{booktabs}
\usepackage{flushend}

\crefname{equation}{}{} % remove Eq. prefix
\Crefname{equation}{Equation}{Equations} % remove Eq. prefix (ok to start sentence with "Equation...")
\Crefname{figure}{Fig.}{Figs.} % capitalized Fig.

\newcommand{\mpcritic}{{\texttt{MPCritic}}\xspace}
\newcommand{\smpcritic}{{\texttt{soft\,MPCritic}}\xspace}

\newcommand{\qfun}{\ensuremath{\mathcal{Q}}}
\newcommand{\vfun}{\ensuremath{\mathcal{V}}}
\newcommand{\ffun}{\ensuremath{\mathcal{F}}}

\title{\LARGE \bf Soft MPCritic: Value-Based Reinforcement Learning for Non-Differentiable  Predictive Control
}

\author{Thomas Banker, Nathan P. Lawrence, and Ali Mesbah% <-this % stops a space
\thanks{The authors are with the 
        Department of Chemical and Biomolecular Engineering, 
        University of California, Berkeley, CA 94720, USA.
        {\tt\small mesbah@berkeley.edu}}%
        \thanks{This work was supported by the U.S. Department of Energy, Office of Science, Office of Fusion Energy Sciences under award DE‐SC0024472.}% <-this % stops a space
}

\begin{document}

\maketitle
\thispagestyle{empty}
\pagestyle{empty}

%%%%%%%%%%%%%%%%%%%%%%%%%%%%%%%%%%%%%%%%%%%%%%%%%%%%%%%%%%%%%%%%%%%%%%%%%%%%%%%%
\begin{abstract}
Reinforcement learning (RL) and model predictive control (MPC) offer complementary strengths, yet existing MPC-RL frameworks commonly rely on gradient-based optimization and therefore cannot handle the non-differentiable dynamics and objectives common in RL.
We propose \smpcritic, an MPC-RL framework that learns a soft value function and uses sampling-based planning both for online control and for generating value targets.
Specifically, \smpcritic instantiates MPC with model predictive path integral control (MPPI) and learns a terminal $\qfun$-function through fitted value iteration.
Because the value targets are computed by the same planner used for control, the learned value function is aligned with the planner and implicitly extends its effective horizon.
To keep target generation tractable, we introduce an amortized warm-start strategy that reuses the open-loop action sequences planned during online control to initialize batched MPPI computations of value targets, reducing computational cost without sacrificing solution quality.
For robustness to model error, \smpcritic plans over scenarios drawn from an ensemble of dynamics models trained for next-step prediction.
On both classical and constrained control tasks, \smpcritic learns effective policies with non-differentiable objectives, particularly where finite-horizon MPPI alone falls short.
\smpcritic thus offers a practical blueprint for synthesizing MPC policies directly from black-box objectives, without differentiable surrogates.
\end{abstract}

%%%%%%%%%%%%%%%%%%%%%%%%%%%%%%%%%%%%%%%%%%%%%%%%%%%%%%%%%%%%%%%%%%%%%%%%%%%%%%%%
\section{Introduction}

% challenge in model-based/mpc approaches is matching the performance, ease, speed of model-free methods
% we seek a plug-in mpc approach that has these benefits plus mpc benefits (safety, robustness--tho not the focus of this work) -- there are lots of model-based approaches with similar goals but we are specifically interested in giving mpc this flexibility

%% design philosophy of rl and mpc; seek to merge them
\Ac{RL} and \ac{MPC} are contrasting strategies for designing decision-making agents.
\Ac{MPC} accounts for system dynamics and constraints to construct a notion of value to be optimized online in a receding-horizon fashion \cite{rawlingsModelPredictiveControl2017}; however, its horizon is finite and computationally costly to increase.
Meanwhile, a core design philosophy driving (actor-critic) \ac{RL} is value-based backup followed by policy extraction \cite{lillicrap2016Continuouscontrol, haarnoja2018Softactorcritic}; but RL can require copious environment interactions.
Models can reduce this burden, for example, by simulating experience \cite{frauenknecht2024Trustmodel}, or by constructing value targets \cite{feinberg2018model} and losses \cite{byravan2020imagined} for policy learning.
However, such approaches can be brittle when they over-rely on model-generated experience \cite{jafferjee2020HallucinatingValue, barkley2024StealingThat}, and enforcing constraints online remains a challenge since the model is not directly used for control.

\ac{RL} and \ac{MPC} offer complementary benefits.
\Ac{RL} can provide adaptivity to an otherwise static \ac{MPC}, targeting infinite-horizon control objectives.
Conversely, \ac{MPC} can replace conventional function approximators (i.e., deep neural networks) in \ac{RL} to enforce constraints and accelerate learning \cite{banker2025local}.
These mutual benefits motivate hybrid frameworks that integrate the principled planning and constraint satisfaction of \ac{MPC} with the versatile reward-driven learning of \ac{RL} \cite{lawrence2025view, reiter2026synthesis}.

%% mpc-rl work; model learning objectives
There is a plethora of ways in which \ac{MPC} and \ac{RL} can be synergized.
Taking a value-based perspective, \ac{RL} can augment \ac{MPC} with a terminal value function to design decisions that target infinite-horizon performance \cite{lawrence2025view}.
This perspective also includes \ac{MPC} providing model-based value targets for training a critic network \cite{bhardwaj2020InformationTheoretic}.
Alternatively, from a policy-based perspective, \ac{MPC} can augment \ac{RL} by refining the actions of a pretrained policy online \cite{qu2024RLDrivenMPPI, wang2025ResidualMPPIOnline}; a strategy that has also been combined with task-oriented model learning, in which the dynamics model is trained to predict reward and value rather than to minimize state-prediction error \cite{hansen2024TDMPC2Scalable}.
Differentiable \ac{MPC} extends this reward-driven paradigm further by treating the controller itself as a parameterized policy, so that its cost and model parameters can be learned end-to-end by differentiating through the optimization \cite{amos2019DifferentiableMPC, gros2020data}.
However, differentiable \ac{MPC} faces several practical challenges: the learned \ac{MPC} problem may become infeasible, solving and differentiating through it at every step of \ac{RL} training is computationally expensive, and the highly specialized software tools of \ac{MPC} and \ac{RL} are difficult to integrate \cite{reiter2026synthesis}.

To mitigate these challenges, \mpcritic \cite{lawrence2025mpcritic} was introduced to enable efficient integration of \ac{MPC} and \ac{RL} through classical dynamic programming principles. 
\mpcritic follows an actor-critic design, wherein a ``fictitious'' controller serves as an approximate \ac{MPC} solution \cite{drgona2024LearningConstrained} to enable the integration of \ac{MPC} within \ac{RL} without explicitly solving \ac{MPC} or computing its sensitivities.
% The design uses the \ac{DPC} method \cite{drgona2024LearningConstrained} in an iterative, online training context.
Meanwhile, \mpcritic preserves a standard \ac{MPC} for accurate planning during online control, where constraint satisfaction is most important.
However, \mpcritic---like most existing \ac{MPC}-\ac{RL} frameworks---requires a differentiable control objective that encodes the system constraints and dynamics model to train the actor.
This differentiability assumption does not always hold in \ac{RL} problems, where rewards or dynamics may be non-differentiable.

Here, we propose \smpcritic, an alternative framework for efficient \ac{MPC}-\ac{RL} integration that operates entirely in value space.
\smpcritic instantiates \ac{MPC} via \ac{MPPI} \cite{williams2016Aggressivedriving, williamsInformationTheoreticMPC2017, honda2025model}, trading optimization fidelity \cite{homburgerOptimalitySuboptimality} for amortized computation and support for non-differentiable control objectives and dynamics.
The policy \emph{implicitly} induced by \ac{MPPI} uses importance sampling to approximate a soft minimum over trajectories, requiring only evaluations of the objective and dynamics, not their derivatives.
This soft minimum corresponds to the free energy of the system, which naturally connects \ac{MPPI} to soft value iteration \cite{haarnojaReinforcementLearning, levine2018reinforcement}; the hard minimization of classical dynamic programming \cite{bertsekas2012dynamic} is replaced by a soft, temperature-weighted counterpart.
Motivated by this connection, we augment the \ac{MPPI} planner with a terminal value function, so that the planner can serve two roles: online control and generating temporal-difference-style targets to learn the value function.
Temporal-difference learning aligns the value function with the planner, effectively extending its horizon for online decision-making.

Nonetheless, generating \ac{MPPI} targets requires solving a sampling-based planning problem at every sampled state in every value-iteration update, which quickly becomes a computational bottleneck.
Building on warm-starting, which is standard practice for reducing online computation in \ac{MPPI}, we reuse previous solutions to keep target generation tractable during \ac{RL} updates. 
Because targets change gradually between value iterations, we warm-start each state's \ac{MPPI} target from its previous solution and store the updated solution in the replay buffer for the following iteration, amortizing target generation across training. The performance of \smpcritic is evaluated on classical and constrained control problems with non-differentiable control objectives.
These case studies demonstrate the importance of a terminal value function for long-horizon planning and of warm-started value targets for computational efficiency.
We also characterize how control performance and constraint satisfaction depend on the planning horizon, the number of \ac{MPPI} samples, and the use of a dynamics model ensemble.

%%%%%%%%%%%%%%%%%%%%%%%%%%%%%%%%%%%%%%%%%%%%%%%%%%%%%%%%%%%%%%%%%%%%%%%%%%%%%%%%
\section{Problem setting}

\subsection{Markov decision processes}

We consider \iac{MDP} comprising an environment and agent.
The environment has states $s \in \state$ that evolve as actions $a \in \action$ are proposed by the agent.
We assume deterministic transitions $s' = f(s,a)$, where $s' \in \state$ denotes the next state, starting from an initial state $s_0 \sim p(s_0)$.

States and actions are scored with a cost function $\ell: \state \times \action \to \reals$.
The agent should design actions that account for their cumulative cost.
As a reference, we first examine the so-called \emph{free energy} of the system. 
The free energy considers an \emph{open-loop} prior $\tilde{\beta}$ over actions and the cost of the resulting trajectories.
We denote the induced distribution over trajectories of horizon $H$ by $\tau \sim p^{\tilde{\beta}}$, where $\tau = \{ s_0, a_0, \ldots, s_{H}, a_{H} \}$.
Note $\tau$ depends only on the initial state $s_0$ and sequences of actions, denoted by $\alpha = \{ a_0, \ldots, a_{H} \}$; hence we write $\tau = \tau(\alpha, s_0)$.
Taking $J$ to be a real-valued function measuring cumulative cost, the free energy at $s_0$ is defined as
\begin{equation}
    \ffun(s_0) = - \lambda \log \left( \EE_{\tau \sim p^{\tilde{\beta}}} \left[ \exp \left( -\frac{1}{\lambda} J(\tau) \right) \right] \right),
\label{eq:free_energy}
\end{equation}
where $\lambda > 0$ is a temperature parameter.
Intuitively, free energy is a soft minimum over random trajectories.
The temperature $\lambda$ influences the sharpness of the soft minimum operator, and $\tilde{\beta}$ determines the scope and texture of the action space.

For any distribution over actions $\beta$, it can be shown (by Jensen's inequality) that 
\begin{equation}
    \ffun(s_0) \leq -\lambda \mathbb{E}_{\tau \sim p^{\beta}} \left[ \log\left( \prod_{t = 0}^{H} \frac{\tilde{\beta}(a_t)}{\beta(a_t)} \exp\left( -\frac{1}{\lambda} J(\tau) \right) \right)  \right].
\label{eq:jensen_bound}
\end{equation}
Given this upper bound, and with a slight abuse of notation, it can be verified that the optimal distribution of open-loop action sequences has density
\begin{equation}
    \beta^\star (\alpha) = \frac{1}{\eta} \exp\left( -\frac{1}{\lambda} J\big(\tau(\alpha,s_0)\big) \right)  \prod_{t=0}^{H}\tilde{\beta}(a_t),
\label{eq:opt_pdf}
\end{equation}
where $\eta$ is a normalizing constant.
While \cref{eq:opt_pdf} gives an elegant solution to the problem, sampling directly from $\beta^\star$ is infeasible.
The goal of the agent is to design a near-optimal control distribution over actions that approximately matches the free energy of the system.
Next, we outline two methodological frameworks for this problem.

\subsection{Model predictive path integral control}
%% mppi

\Ac{MPPI} approximately samples from $\beta^\star$ through importance sampling and Monte Carlo estimation \cite{williams2016Aggressivedriving, williamsInformationTheoreticMPC2017, homburgerOptimalitySuboptimality}.
\Ac{MPPI} aims to find optimal control inputs that minimize the \ac{KL} divergence between the optimal distribution $\beta^\star$ and the design distribution $\beta$.
The principal result is that the optimal control inputs can be expressed as an expectation of weighted actions over $\beta$
\begin{equation}
    u_t^\star = \frac{1}{\eta} \mathbb{E}_{\alpha \sim \beta(\alpha)} \left[ w(\alpha) a_t \right].
\end{equation}
Assuming, as a standard example, $\tilde{\beta} = \mathcal{N}(0, \Sigma)$ and $\beta = \mathcal{N}(u, \Sigma)$, where $u$ is a commanded control input,
actions are sampled through the change of variables $a = u + \epsilon$.
Sampling $N$ noise sequences $\varepsilon^{(n)} = \{ \epsilon_0^{(n)}, \ldots, \epsilon_{H}^{(n)} \}$, the (unnormalized) weights at $s_0$ are given by
\begin{multline}
w(\varepsilon^{(n)}) = \exp\bigg( -\frac{1}{\lambda} \Big( J\big(\tau(\upsilon + \varepsilon^{(n)}, s_0)\big) \\+ \frac{\lambda}{2} \sum_{t=0}^{H} u_t\transpose \Sigma^{-1} \big( u_t + 2 \epsilon_t^{(n)} \big)\Big)\bigg),
\end{multline}
where $\upsilon = \{ u_0, \ldots, u_{H} \}$
% Then we have 
% $w(\alpha^{(n)}) = \frac{1}{\eta} \hat{w}(\varepsilon^{(n)})$, where
and the (approximate) normalization constant in \cref{eq:opt_pdf} is given by
\begin{equation}
    \eta = \frac{1}{N}\sum_{n=1}^{N} w(\varepsilon^{(n)}),
\end{equation}
with $-\lambda\log(\eta)$ estimating the free energy $\ffun(s_0)$ in \cref{eq:free_energy}.

This weighted structure highlights low-cost trajectories and ignores ``bad'' ones.
Control inputs are then updated iteratively
\begin{equation}
    u^{(i+1)}_t = u^{(i)}_t + \frac{1}{\eta N} \sum_{n=1}^{N} w (\varepsilon^{(n)})  \epsilon_t^{(n)},
\label{eq:u_iter}
\end{equation}
returning solution $\upsilon^+$ after all iterations.
\Ac{MPPI} is applied in a receding-horizon fashion: the online policy applies $u^+_0$ and repeats the optimization at the next state, warm-started with the shifted solution
\begin{equation}
    \upsilon = \mathcal{T}(\upsilon^+) = \{ u_1^+, u_2^+, \ldots, u_{H}^+, u_\text{init} \},
\label{eq:warm-start}
\end{equation}
where $\mathcal{T}$ shifts $\upsilon^+$ and appends the initial guess $u_\text{init}$.

\subsection{Soft value functions}
%% soft value iteration

The free energy \cref{eq:free_energy} and \ac{MPPI} consider open-loop, finite-horizon trajectories.
To reason about closed-loop, infinite-horizon behavior, we follow soft value iteration \cite{haarnojaReinforcementLearning, levine2018reinforcement} and define value functions recursively.
The \emph{soft action-value function} and \emph{soft value function} are the fixed point of
\begin{align}
    \qfun(s,a) &= \ell(s,a) + \gamma\vfun\big(f(s,a)\big),
    \label{eq:soft_q_value}\\
    \vfun(s) &= -\lambda \log \left( \mathbb{E}_{a \sim \tilde{\beta}}\left[ \exp\left( -\frac{1}{\lambda}\qfun(s,a) \right) \right] \right),
    \label{eq:soft_v_value}
\end{align}
where $0 < \gamma < 1$ is a discount factor.
Substituting \cref{eq:soft_q_value} into \cref{eq:soft_v_value} gives a \emph{soft Bellman optimality equation} in $\qfun$ alone.
While $\qfun$ and $\vfun$ are generally unknown, the resulting recursion enables iterative solution methods that decompose the infinite-horizon objective into recursive subproblems \cite{bertsekas1996neuro,bertsekas2012dynamic}, and its approximations underpin modern \ac{RL} algorithms like \ac{SAC} \cite{haarnoja2018Softactorcritic}.

\Cref{eq:soft_v_value} is the one-step analogue of \cref{eq:free_energy}, with all future cost summarized by $\qfun$.
Hence, $\qfun$ can be taken as an objective function to be (softly) optimized, for example, using the \ac{MPPI} approach.
Proceeding on this path, with $J = \qfun$ and $\alpha = a$, we arrive at the optimal policy
\begin{align}
\pp{\pi^\star}{a}{s} &= \frac{1}{\eta} \exp\left( -\frac{1}{\lambda} \qfun (s,a)  \right) \tilde{\beta}(a) \\
&\propto \exp\left( -\frac{1}{\lambda} \left( \qfun(s,a) - \vfun(s) \right)  - \frac{1}{2} \norm{a}_{\Sigma^{-1}}^2 \right),
\end{align}
with $\vfun(s) = -\lambda \log(\eta)$.
Thus $\pi^\star$ balances the advantage of each action over $\vfun$ against control effort.

In the undiscounted case, unrolling \cref{eq:soft_q_value,eq:soft_v_value} over a finite horizon $H$ collapses them into the form of \cref{eq:free_energy}: a free energy with the value-augmented cost $J(\tau) = \sum_{t=0}^{H-1}\ell(s_t,a_t) + \qfun(s_H,a_H)$.
This form for $\vfun$ is attractive: it can be estimated by \ac{MPPI} through sampling, while $\qfun$ accounts for all cost beyond the horizon.
Motivated by this identity, we define the $H$-step soft value as the free energy with discounted, value-augmented cost % the (approximate) $H$-step soft value
\begin{multline}
  \vfun^\text{MPPI}(s) = -\lambda \log \bigg( \EE_{\tau \sim p^{\tilde{\beta}}} \bigg[ \exp\Big( -\tfrac{1}{\lambda} \big( \textstyle\sum_{t=0}^{H-1} \gamma^t\ell(s_t,a_t) \\
  + \gamma^H\qfun(s_H,a_H) \big) \Big) \,\bigg|\, s_0 = s \bigg] \bigg),
\label{eq:hstep_value}
\end{multline}
which is exact for $H = 0$ or in the undiscounted case and upper-bounds $\vfun(s)$ otherwise, since discounting effectively lowers the temperature at later steps of the recursion; the gap vanishes as $\gamma \to 1$.
We use \cref{eq:hstep_value} in two roles: to design actions via \ac{MPPI} that account for long-term cost, and to provide $H$-step targets that accelerate the learning of $\qfun$.
The $H$-step lookahead refines value estimates from an imperfect $\qfun$ with model-based rollouts.
Further, the use of the same $H$-step soft value for both roles aligns $\qfun$ with the planner that uses it.

%%%%%%%%%%%%%%%%%%%%%%%%%%%%%%%%%%%%%%%%%%%%%%%%%%%%%%%%%%%%%%%%%%%%%%%%%%%%%%%%
\section{\smpcritic framework}

\begin{figure}[t!]
    \centering
    \includegraphics[width=1.\linewidth]{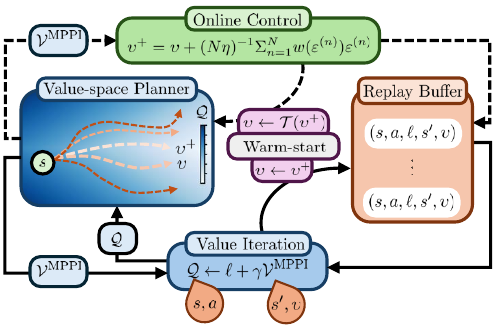}
    \caption{
    \smpcritic exists entirely in value space, combining soft value functions and \ac{MPPI} for autonomous self-improvement.
    During online control, the value-space planner updates the initial open-loop actions $\upsilon$ to align with the soft minimum of the $\qfun$-function.
    The updated sequence $\upsilon^+$, along with transition $(s,a,\ell,s')$, is stored in the replay buffer to warm-start the planner's value targets $\vfun^{\text{MPPI}}$.
    After a $\qfun$-iteration step, the planner is updated with the new $\qfun$, and the further-refined solution $\upsilon^+$ is stored for later reuse.
    }
    \label{fig:concept}
\end{figure}

\smpcritic is a value-based framework for efficiently integrating \ac{MPC} and \ac{RL}.
To accommodate non-differentiable objectives, constraints, and dynamics that commonly occur in a wide range of decision-making problems, \smpcritic instantiates \ac{MPC} via \ac{MPPI}.
Drawing on the connection between \ac{MPPI} and soft value iteration, a value-augmented \ac{MPPI} planner performs online control and generates targets for learning its own terminal $\qfun$-function, aligning the planner and $\qfun$-function (\cref{fig:concept}).
To keep the approach computationally tractable, target generation is amortized over the course of training via warm-starting.
The key components of \smpcritic are discussed below.

\subsection{Fitted $\qfun$-iteration via MPPI}

The value architecture used in \smpcritic is
\begin{equation}    
	\vfun^{\text{MPPI}} (s) \approx -\lambda \log \left( \mathbb{E}_{\tau_\psi \sim p^{\tilde{\beta}}} \left[ \exp\left( - \frac{1}{\lambda} J_\phi(\tau_\psi) \right) \middle| s_0 = s \right] \right),
\label{eq:mppi_value}
\end{equation}
where $J_\phi(\tau_\psi) = \sum_{t=0}^{H-1} \gamma^t  \ell(s_t, a_t) + \gamma^{H} \qfun_\phi(s_H, a_H)$ and $\tau_\psi$ refers to trajectories under a learnable model $f_\psi$, with $\psi$ parameterizing $f_\psi$ and $\phi$ parameterizing $\qfun_\phi$.
$f_\psi$ is trained to minimize the \ac{MSE} between its prediction and the observed next state; however, other objectives or multi-step targets may be used.
Similar to \cite{chua2018deep}, uncertainty in $\psi$ can be mitigated by training an ensemble of dynamics models.
\Ac{MPPI} therefore considers a branching dynamics model of the form
\begin{align}
    \psi &\sim p(\psi)\\
    s' &= f_\psi (s,a),
\end{align}
where models $\psi$ are sampled uniformly.
Concurrently, $\qfun_\phi$ is trained to approximately solve the soft Bellman equation such that $\qfun_\phi \approx \qfun$, i.e., the fixed point of \cref{eq:soft_q_value,eq:soft_v_value}.
Fitted $\qfun$-iteration is an iterative technique for training $\qfun_\phi$; it was introduced by \cite{riedmiller2005neural} and made tractable in high-dimensional continuous spaces by \cite{lillicrap2016Continuouscontrol}.
Here, we use it as an umbrella term, as it is a prevalent idea and this work is compatible with the stabilization techniques introduced since the nominal algorithm of \cite{riedmiller2005neural}. Additionally, we note that \eqref{eq:mppi_value} is an idealization; in practice, its expectation is approximated by a finite number of importance-weighted samples, and both the dynamics model and the $\qfun$-function are learned approximations.

% \begin{remark}
%     \Cref{eq:mppi_value} is an idealization; in practice, its expectation is approximated by a finite number of importance-weighted samples, and both the dynamics model and the $\qfun$-function are learned approximations.
% \end{remark}

We define the objective
\begin{equation}
    \mathcal{L}_\qfun = \text{MSE}\left( \qfun_\phi, \ell + \gamma \vfun^{\text{MPPI}} \right),
\label{eq:qfun_obj}
\end{equation}
where the \ac{MSE} is computed over a batch from the replay buffer $\mathcal{D}$ containing tuples of the form $(s,a,\ell,s',\upsilon)$.
$\ell$ is the observed cost at $s,a$ and $\vfun^{\text{MPPI}}$ is the \ac{MPPI} value in \cref{eq:mppi_value} evaluated at $s'$.
\Cref{eq:qfun_obj} is only a nominal objective, as variations involving double or target $\qfun$-networks are possible \cite{lillicrap2016Continuouscontrol, haarnoja2018Softactorcritic}.
If $H=0$ in $\vfun^{\text{MPPI}}$, then \cref{eq:qfun_obj} will be a soft version of classical fitted $\qfun$-iteration powered by \ac{MPPI}, marking a conceptual similarity to \ac{SAC}.
Taking $H>0$ embeds short-horizon model information into the targets.
This same \ac{MPPI} is used for online control, making $\qfun$ aligned with the model-based \ac{MPPI} value estimate and implicitly extending its prediction horizon.
Without differentiating the \ac{MPPI} targets, the weights $\phi$ are updated iteratively using some form of gradient descent as new data is added to the replay buffer.
$\upsilon$ is included in the replay tuples to make the \ac{MPPI} target computationally efficient.

\subsection{Amortized \smpcritic}
% discuss the central axes/dimensionality bottlenecks

 \begin{algorithm}[tb]
 \caption{Value-augmented \ac{MPPI} with \smpcritic}
 \begin{algorithmic}[1]
 \renewcommand{\algorithmicrequire}{\textbf{Input:}}
 \renewcommand{\algorithmicensure}{\textbf{Output:}}
 \REQUIRE $s_0, \upsilon, p(\psi), \textcolor{mpcritic}{\phi}$
 \FOR {each \ac{MPPI} iteration}
 \FOR {$n\in[1, 2, \ldots, N]$}
 \FOR {$t\in[0, 1, \ldots, H]$}
 \STATE $\epsilon_t^{(n)} \sim \distNorm(0,\Sigma)$
 \STATE $a_t \leftarrow u_t + \epsilon_t^{(n)}$
 \STATE $\psi \sim p(\psi)$
 \STATE $s_{t+1} \leftarrow f_{\psi}(s_t, a_t)$
 \ENDFOR
 \STATE $J \leftarrow \sum_{t=0}^{H-1} \gamma^t \ell(s_t, a_t) + \textcolor{mpcritic}{\gamma^H \qfun_\phi( s_H, a_H )}$
 \STATE $w^{(n)} \leftarrow \exp( -\frac{1}{\lambda} ( J + \frac{\lambda}{2} \sum_{t=0}^{\textcolor{mpcritic}{H}} u_t\transpose \Sigma^{-1} ( u_t + 2 \epsilon_t^{(n)})))$
 \ENDFOR
 \STATE $\eta \leftarrow \frac{1}{N} \sum_{n=1}^N w^{(n)}$
 \FOR {$t\in[0, 1, \ldots, H]$}
 \STATE $u_t \leftarrow u_t + \frac{1}{\eta N} \sum_{n=1}^N w^{(n)} \epsilon_t^{(n)}$
 \ENDFOR
 \ENDFOR
  \STATE $\upsilon^+ \leftarrow \upsilon$
 \STATE $\vfun^{\text{MPPI}} \leftarrow -\lambda \log(\eta)$
 \RETURN $\upsilon^+, \vfun^{\text{MPPI}}$
 \end{algorithmic} 
 \label{alg:mppi}
 \end{algorithm}

The number of trajectory samples is a potential bottleneck for any framework involving \ac{MPPI}, such as \smpcritic.
Estimation of the optimal control inputs quickly becomes infeasible as $H \rightarrow\infty$, meaning the number of trajectories is coupled with the prediction horizon.
\smpcritic addresses this by fixing the prediction horizon and caching long-horizon planning with a terminal \qfun-function, i.e.,
\cref{eq:mppi_value}.
This enables action design that accounts for long-term cost with a tractable number of planning trajectories, targeting global optimality in the soft Bellman sense.
\Cref{alg:mppi} is a generic instantiation of this, which can be expanded to incorporate covariance and/or temperature annealing \cite{homburgerOptimalitySuboptimality}, baseline policies \cite{wang2025ResidualMPPIOnline,qu2024RLDrivenMPPI}, non-Gaussian distributions, and other recent advancements in \ac{MPPI} \cite{honda2025model}.

While a terminal \qfun-function can alleviate sampling challenges during online control, it shifts the burden of computation to $\qfun$-iteration.
Furthermore, for \cref{alg:mppi} to be successful online, $\qfun_\phi$ should be aligned with the \ac{MPPI} control objective.
Hence, we use \ac{MPPI} for both online control and computing $\qfun$-iteration targets.
However, considering a fixed number of trajectory samples for computing targets, the cost of \ac{RL} scales with the number of $\qfun$-iteration steps.
\smpcritic addresses this issue by amortizing target computations at each $\qfun$-iteration step.
Recall \cref{eq:u_iter} is an iterative update rule.
Thus, the \ac{MPPI} optimization for \ac{RL} updates need not start from scratch; open-loop controls $\upsilon$ calculated online can be recycled for target computation.
This inspires the proposed warm-starting strategy for target computations, exemplified in \cref{alg:warmstart RL}.

The benefits of warm-starting targets as in \cref{alg:warmstart RL} include: 1) fewer trajectory samples needed for \ac{RL} updates; and 2) targets are iteratively refined as learning proceeds.
Regarding the former, online control demands a sizable number of trajectories to acquire near-optimal solutions.
Hence, when applied to target initialization, only minor refinement is typically necessary.
These refinements occur each time a tuple $(s, a, \ell, s', \upsilon)$ is sampled from the replay buffer, meaning $\upsilon$ adapts with the model and terminal $\qfun$-function throughout training.%, which can be accomplished in significantly fewer trajectory samples than that used online 

 \begin{algorithm}[tb]
 \caption{$\qfun$-iteration with amortized \smpcritic}
 \begin{algorithmic}[1]
 \renewcommand{\algorithmicrequire}{\textbf{Input:}}
 \renewcommand{\algorithmicensure}{\textbf{Output:}}
% \REQUIRE in
 \STATE Initialize $\upsilon, \phi, \psi$
  \FOR {each environment step}
  \STATE $\upsilon^+, \cdot \leftarrow \text{MPPI}(s, \upsilon, p(\psi), \phi) \hfill\triangleright\ \text{\cref{alg:mppi}}$
  \STATE $a \leftarrow u_0^+$
  \STATE $s' \leftarrow f(s,a)$, \ $\ell \leftarrow \ell(s,a)$
  \STATE $\upsilon \leftarrow \mathcal{T}(\upsilon^+)$
  \STATE $\mathcal{D} \leftarrow \mathcal{D} \cup \{(s,a,\ell,s',\upsilon)\}$
  \FOR {each update step}
  \STATE $\mathcal{B} = \{(s,a,\ell,s',\textcolor{warmstart}{\upsilon})\} \sim \mathcal{D}$
  \FOR {each tuple in $\mathcal{B}$}
    \STATE $\textcolor{warmstart}{\upsilon^+}, \vfun^{\text{MPPI}}(s') \leftarrow \text{MPPI}(s', \textcolor{warmstart}{\upsilon}, p(\psi), \phi)$
    \STATE Replace $\textcolor{warmstart}{\upsilon}$ with $\textcolor{warmstart}{\upsilon^+}$ in $\mathcal{D}$
  \ENDFOR
  \STATE $\phi \leftarrow \phi - \kappa \nabla \mathcal{L}_{\qfun} \hfill\triangleright\ \text{e.g., \cref{eq:qfun_obj}}$
  \STATE $\psi \leftarrow \psi - \kappa \nabla \mathcal{L}_{f} \hfill\triangleright\ \text{e.g., \acs{MSE}}$\label{algeq:ID}
  \ENDFOR 
  \ENDFOR
 \end{algorithmic} 
 \label{alg:warmstart RL}
 \end{algorithm}
 
Empirically, we observe that warm-starting \ac{MPPI} targets can be highly effective; with only $\approx 10 \%$ of samples used online, significantly reducing the computational cost of the framework.
This can be explained through the weighted structure behind \cref{eq:u_iter}.
Consider two samples $\varepsilon^{(j)}$ and $\varepsilon^{(i)}$, giving the ratio of weights
\begin{multline}
    \frac{w(\varepsilon^{(j)})}{w(\varepsilon^{(i)})} = \exp \bigg( -\frac{1}{\lambda} \Big( J\big(\tau(\upsilon + \varepsilon^{(j)}, s_0)\big) \\ - J\big(\tau(\upsilon + \varepsilon^{(i)}, s_0)\big) \Big)
    - \textstyle\sum_{t=0}^{H} u_t\transpose \Sigma^{-1} \big( \epsilon_t^{(j)} - \epsilon_t^{(i)} \big) \bigg).
\end{multline} 
The weight ratio ensures that even a linear increase in cost $J$ results in an exponential decrease in influence when updating $\upsilon$.
While not used in the ensuing case studies, the risk of all samples increasing the cost---thereby degrading $\upsilon$---can be mitigated without increasing sample size by annealing the covariance.
Furthermore, as part of the latter benefit, any suboptimal updates are naturally corrected in subsequent iterations, as previous target solutions are recycled to warm-start the optimization.
Reusing target solutions in subsequent iterations also helps prevent $\upsilon$ from becoming stale as $\qfun_\phi$ is continually updated.
Thus, amortization via warm-started targets yields substantial computational savings without significantly compromising the overall framework's performance.

\subsection{Broader scope of \smpcritic}

\smpcritic lives in soft value space; however, it could also be adapted to a traditional \ac{DP} setting.
This can be accomplished by targeting the usual Bellman optimality equation.
Further, the resulting value function-augmented \ac{MPC} would directly minimize cost under the system model.
Nonetheless, \ac{MPPI} is useful in the \smpcritic framework for several reasons.
The model is learned online in tandem with the $\qfun$-function, meaning targets that minimize the predicted cost exactly may be detrimental to overall stability and performance over the course of training \cite{fujimoto2018addressing}.
Additionally, \ac{MPPI} requires only forward evaluations of the dynamics, cost, and value models, with no gradients or structural assumptions, which eases software integration, a key bottleneck in the \ac{MPC}-\ac{RL} interface \cite{reiter2026synthesis}.
Therefore, while \ac{MPPI} is suboptimal relative to a deterministic optimal control problem \cite{homburgerOptimalitySuboptimality}, its benefits are worthwhile.

%%%%%%%%%%%%%%%%%%%%%%%%%%%%%%%%%%%%%%%%%%%%%%%%%%%%%%%%%%%%%%%%%%%%%%%%%%%%%%%%
\section{Case studies}
\label{sec:examples}

\smpcritic is tested on two case studies: the first examines how amortization enables tractable learning, and the second builds upon this to identify the critical components of \smpcritic.
The first case study is balancing a double inverted pendulum, a classical control environment.
With a one-dimensional control input, this environment provides a practical testbed for studying the computational benefits of amortizing \qfun-iteration. %, where we measuring compute time and closed-loop cost as a proxy for solution quality.
The computational savings from amortization help \smpcritic scale to more challenging environments with larger action spaces and constraints that conflict with the control goal.
The second case study is a quadrotor with two-dimensional control inputs, requiring planning over multiple constraints encoded in a non-differentiable cost.
Here, we establish the importance of a terminal $\qfun$-function and modeling uncertainty in the dynamics.
We also measure the sensitivity of \smpcritic to the horizon and number of \ac{MPPI} samples, and compare against standard \ac{RL} and \ac{MPPI} baselines.\footnote{Code with full experimental details is available at \url{https://github.com/NPLawrence/soft-mpcritic}}

\subsection{Balancing a double inverted pendulum}

% We first evaluate online learning with \smpcritic on balancing a double inverted pendulum.
We use the standard MuJoCo benchmark environment \cite{towers2024gymnasium} with unknown dynamics but known, discontinuous cost
\begin{equation}
  \ell(s,a) = \lVert z \rVert_W^2 + 
  \begin{cases}
    -10 & y > 1, \\
    0   & \text{otherwise},
  \end{cases}
\end{equation}
where $s \in \mathbb{R}^9$, $a \in [-1,1]$, $z = (x, y-2, \omega_1, \omega_2)$, $(x,y)$ is the pendulum tip position, $\omega_1,\omega_2$ are the hinge angular velocities, and $W = \operatorname{diag}(0.01,\,1,\,0.001,\,0.005)$. We use this benchmark to study how warm-starting makes combining \ac{MPPI} with $\qfun$-iteration tractable, before moving to harder environments that need more $\qfun$-iterations.
Both $\qfun_\phi$ and $f_\psi$ are parameterized as neural networks in \smpcritic, with $4$ networks in the dynamics ensemble.
Action sequences are designed over a horizon $H=5$ with $\Sigma=0.1$, $\lambda=0.1$, and $\gamma=0.99$ using $N=600$ and $N_\text{target}=60$ \ac{MPPI} samples for control and value generation, respectively.
% We analyze how warm-starting and an ensemble of dynamics models affect computation and performance.

\begin{figure}[tb]
    \centering
    \includegraphics[width=0.95\linewidth]{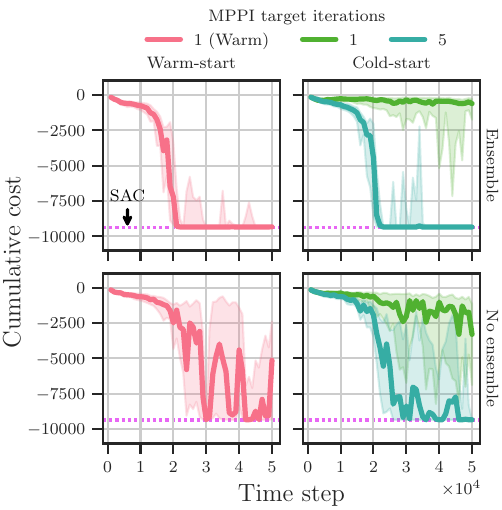}
    \caption{
    Warm- vs.\ cold-starting \ac{MPPI} targets with a single dynamics model or an ensemble.
    Lines represent the median of $10$ seeds, shading the interquartile range.
    The dashed line marks \ac{SAC}'s median after $10^6$ time steps.
    }
    \label{fig:dip}
\end{figure}

Before measuring computation times, we check that warm-starting still enables high-performance control; \cref{fig:dip} shows the performance of four learning configurations for \smpcritic.
% The grid of plots in \cref{fig:dip} displays the performance of four learning configurations for \smpcritic.
With warm-starting and one \ac{MPPI} target iteration (top-left), \smpcritic quickly minimizes the cost, approximately matching that of \ac{SAC}.
In contrast, with cold-started targets and one iteration (top-right), learning to control the system is a challenge.
% This is in stark contrast with cold starting \ac{MPPI} for computing \qfun-function targets (top-right).
% Cold starting generally increases variance during learning but, if given enough iterations, can achieve similarly performing solutions.
However, if given enough iterations, cold-starting can achieve similarly performing solutions at the expense of additional compute.
% However, this comes with additional compute, whereas warm-starting can recycle the online solution and continually refine it.
Warm-starting instead recycles the online solution and continually refines it.
The open-loop controls $\upsilon$ stored in the replay buffer are iteratively updated to best approximate the optimal sequence for the current $\qfun_\phi$ and $f_\psi$.
Hence, in this example, only a single iteration is necessary to (approximately) minimize cost.

Furthermore, using a single dynamics model can result in suboptimal solutions when warm- (bottom-left) or cold-starting (bottom-right).
This can be attributed to compounding prediction errors over the planning horizon, which manifests in both learning the $\qfun$-function and online control to ultimately hinder performance.
In contrast, planning over an ensemble provides a degree of robustness to parametric uncertainties, meaning \ac{MPPI} favors low-cost trajectories with high agreement among the ensemble.
As seen in \cref{fig:dip}, this can aid in tightening the range of costs when warm- or cold-starting.
We revisit this point in the next case study, where accurate planning is key for reliable constraint satisfaction.

Since, warm-starting can achieve comparable performance to cold-starting, we analyze its utility in amortizing the computational cost of \cref{alg:warmstart RL}.
\Cref{tab:sps} reports the environment steps per second for each configuration on an Apple M3 Pro 11-core laptop.
As the number of \ac{MPPI} samples used for target computation increases, all approaches experience a degree of slowdown.
At one iteration, warm- and cold-starting are nearly identical in speed; warm-starting costs nothing extra per iteration.
However, recalling \cref{fig:dip}, cold-starting requires multiple \ac{MPPI} iterations to achieve comparable performance to one iteration when warm-starting.
Requiring additional iterations brings the unfavorable scaling of cold-starting to the forefront, increasing wall time by roughly $1.7$--$3.5\times$, depending on the desired solution quality.
Hence, we only consider warm-starting target computation for all further experiments for its computational efficiency with minimal sacrifice to performance.

\begin{table}[tb]
    \normalsize
    \centering
    \caption{Mean and standard deviation of environment steps per second (SPS) during online learning for varying \ac{MPPI} samples and iterations for computing targets; higher is better.}
    \label{tab:sps}
     \begin{tabular*}{\linewidth}{@{\extracolsep{\fill}}ccccc}
        \toprule[1.25pt]
        \multicolumn{1}{c}{} & \multicolumn{4}{c}{MPPI target computation iterations (SPS)} \\
        \cmidrule(l){2-5}
        \multicolumn{1}{c}{} & \multicolumn{1}{c}{Warm} & \multicolumn{3}{c}{Cold} \\
        \cmidrule(rl){2-2} \cmidrule(l){3-5}
            $N_\text{target}$ & $1$ iter. & $1$ iter. & $3$ iters. & $5$ iters. \\
        \midrule
        20 & 184$\pm$2.80 & 181$\pm$4.69 & 110$\pm$1.71 & 78$\pm$0.95 \\
        60 & 124$\pm$2.92 & 125$\pm$3.55 & 61$\pm$0.66 & 40$\pm$0.39 \\
        100 & 95$\pm$1.92 & 95$\pm$1.25 & 42$\pm$0.76 & 27$\pm$0.21 \\
        \bottomrule[1.25pt]
    \end{tabular*}
\end{table}

\subsection{Constrained quadrotor control}

We consider \texttt{safe-control-gym}'s planar quadrotor \cite{yuan2021safecontrolgym} with unknown dynamics, where \smpcritic must plan over a known but non-differentiable cost.
Here, $s = (x, z, \theta, \dot{x}, \dot{z}, \dot{\theta}) \in \mathbb{R}^6$ comprises positions [m], pitch [rad], and associated time derivatives ([m/s] and [rad/s]), while actions $a \in [-1,1]^2$ are normalized motor thrusts.
The agent should continuously lap an annular goal zone $\mathcal{G}$, maximizing its angular velocity $\dot\vartheta$ about the center of the annulus while avoiding obstacles $\mathcal{O}$ and respecting state limits, where
\begin{align}
  \mathcal{O} &= \textstyle\bigcup_{i=1}^{4} \{ s \mid (x-x_i^{\mathcal{O}})^2 + (z-z_i^{\mathcal{O}})^2 \leq 0.04^2 \}, \\
  \mathcal{G} &= \{ s \mid 0.2^2 \leq x^2 + z^2 \leq 0.3^2 \} \setminus \mathcal{O},
\end{align}
with obstacle centers $(x_i^{\mathcal{O}},z_i^{\mathcal{O}})$.
This is encoded by the discontinuous cost
\begin{equation}
  \ell(s,a) = c_1\mathbf{1}\transpose[C\abs{s}-b]_+ +
  \begin{cases}
    -c_2[\dot\vartheta]_+        & s\in\mathcal{G}, \\
    1 - c_2[\dot\vartheta]_-    & s\in\mathcal{O}, \\
    0.1 - c_2[\dot\vartheta]_-  & \text{otherwise},
  \end{cases}
\end{equation}
where $c_1=20$, $c_2=1200$, $C=\operatorname{diag}(0, 0, 1, 1, 1, 0)$ selects the constrained states, $b=(0, 0, 0.15, 0.5, 0.5, 0)$ their limits, $[\cdot]_+ = \max(0,\cdot)$, and $[\cdot]_- = \min(0,\cdot)$.
Therefore, the agent incurs nonnegative cost unless it laps $\mathcal{G}$ within the constraints, i.e., inside $\mathcal{G}$ with $C\abs{s} \leq b$.

Because $\ell$ jumps at the boundaries of $\mathcal{G}$ and $\mathcal{O}$, its gradient carries no information about where these regions lie, so approaches relying on gradient-based optimization such as \mpcritic would require a differentiable surrogate for $\ell$.
Since \smpcritic plans and learns from cost evaluations alone, it accommodates such costs directly.
As in the previous case study, all learnable components of \smpcritic are parameterized as neural networks, with $4$ networks in the dynamics ensemble, $\lambda=0.1$, and $\gamma=0.99$.
Action sequences are designed over a horizon $H=20$ using $N=1000$ and $N_\text{target}=100$ \ac{MPPI} samples for control and value generation, respectively.
For precise control, we found it beneficial to correlate the sampled motor thrusts, using $\Sigma$ with diagonal entries $0.275$ and off-diagonal entries $0.225$, so that sampling perturbations induce smaller torques.

\begin{figure}[tb]
    \centering
    \includegraphics[width=0.95\linewidth]{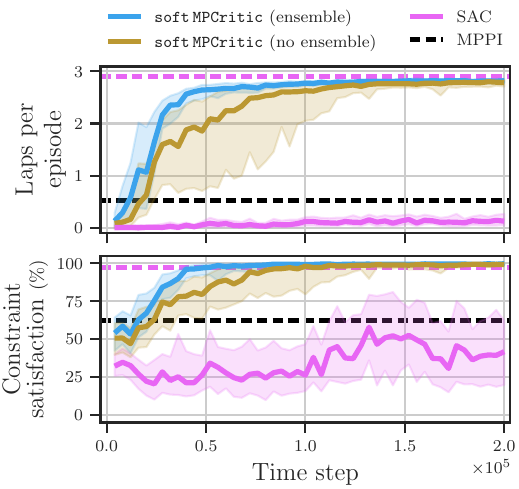}
    \caption{
    Performance of \smpcritic with a single model or an ensemble of dynamics models, compared with \ac{MPPI} using oracle dynamics and \ac{SAC}.
    Lines show the median over $10$ seeds with the interquartile range shaded; \ac{MPPI} is shown as the median over $10^5$ time steps in closed-loop.
    The purple dashed line marks \ac{SAC}'s median after $2\times10^6$ time steps.
    }
    \label{fig:method}
\end{figure}

\Cref{fig:method} isolates two essential ingredients of \smpcritic: the terminal \qfun-function and an ensemble of dynamics models.
Using the same settings as \smpcritic, finite-horizon \ac{MPPI} cannot reliably complete laps within the constraints even with oracle dynamics.
Longer horizons can help, but require more \ac{MPPI} samples.
\smpcritic addresses this limitation with a terminal \qfun-function, which implicitly extends the planning horizon and enables reliable lap completion with no additional online sampling.
The $H$-step model-based targets also accelerate value learning: \smpcritic reaches this performance within $2\times10^5$ time steps, whereas model-free \ac{SAC} requires significantly more.
However, value learning remains sensitive to model error: with a single imperfect dynamics model, $\qfun_\phi$ is learned and optimized under biased predictions, which degrades performance.
The ensemble mitigates this by propagating dynamics uncertainty into both value learning and planning.

\begin{figure}[tb]
    \centering
    \includegraphics[width=0.95\linewidth]{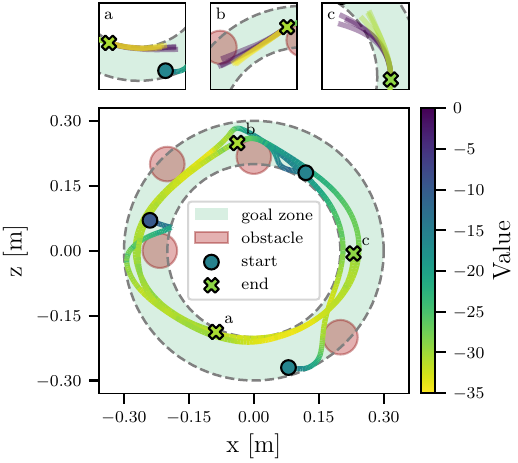}
    \caption{
    Closed-loop \smpcritic trajectories from challenging initial conditions, colored by $\qfun_\phi(s,a)$ (lower is better).
    Insets: trajectory predictions for action sequences randomly sampled via \ac{MPPI} from each end state, colored by their importance-weighted cost.
    }
    \label{fig:traj}
\end{figure}

After learning, we evaluate \smpcritic in closed-loop from challenging states that would violate the constraints with no control (\cref{fig:traj}).
The learned \qfun-function correctly identifies each initial state as high cost.
Furthermore, the agent navigates around the obstacles, remains in the goal zone, and satisfies all other constraints.
This can be attributed to the value-augmented \ac{MPPI} used for online planning over the constraints.
Model-based rollout predictions from the final states for randomly sampled \ac{MPPI} action sequences are shown in each inset.
Because \ac{MPPI} samples are weighted exponentially by cost, those bound to violate constraints or leave the goal zone receive negligible weight when designing control inputs.

\begin{figure}[tb]
    \centering
    \includegraphics[width=0.95\linewidth]{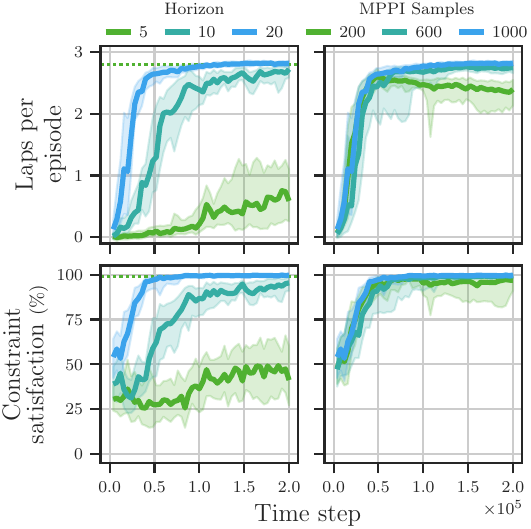}
    \caption{
    Sensitivity of \smpcritic to the horizon (left, $1000$ \ac{MPPI} samples) and \ac{MPPI} samples (right, horizon of $20$).
    Lines show the median over $10$ seeds, with the interquartile range shaded; the dashed line marks the median for a horizon of $5$ after $2\times10^6$ time steps.
    }
    \label{fig:scaling}
\end{figure}

Having shown that \smpcritic achieves high-performance control, we next examine its sensitivity to the horizon $H$ and number of \ac{MPPI} samples $N$.
Starting from $H=20$ and $N=1000$, as used in \cref{fig:method,fig:traj}, we reduce each independently to isolate its effect (\cref{fig:scaling}).
Shortening $H$ slows learning (left), since truncating the model-based targets in \cref{eq:mppi_value} forgoes model information that accelerates value learning when the dynamics models are accurate.
Nonetheless, short horizons still reach high performance with more $\qfun$-iterations, e.g., $H=5$ after $2\times10^6$ time steps.
Reducing $N$ also degrades performance (right).
Because \ac{MPPI} serves both online control and value learning, fewer \ac{MPPI} samples reduce the accuracy of \cref{eq:u_iter} in approximating the optimal controls and increase the variance of the targets in \cref{eq:mppi_value}.
As the number of \ac{MPPI} samples needed to approximate either grows with $H$, $N$ and $H$ are coupled; the strengths of \smpcritic are thus most pronounced when provided sufficient samples for a moderate planning horizon.

\section{Conclusions}

\smpcritic is a value-based framework that integrates soft value iteration with \ac{MPPI}, using a learned terminal value function to extend the planner's effective horizon.
The result is an adaptive model-based planner that learns to optimize long-term performance even when the objective, constraints, or dynamics are non-differentiable; a setting that challenges \ac{MPC}-\ac{RL} frameworks built on gradient-based optimization.
We identify the key design choices that make this integration tractable, including how the value function is trained from \ac{MPPI} rollouts and how it is incorporated as a terminal cost, and demonstrate the approach on classical and constrained control problems.

A natural next step is to extend \smpcritic to high-dimensional action spaces and partially observed settings.
These settings can challenge \ac{MPPI} and \ac{RL}, respectively. %, and thus \smpcritic.
% Improvements could draw on techniques from \ac{MPPI}, such as reverse-KL updates for mode-seeking behavior or covariance annealing for precise control; from \ac{RL}, such as temperature auto-tuning for exploration or using multiple value function approximators for targets; and from control theory, such as joint estimation and control.
Extensions to these settings could draw on \ac{MPPI} techniques, such as reverse-\ac{KL} updates for mode-seeking behavior, or covariance annealing for precise control.
From \ac{RL}, temperature auto-tuning could aid exploration, and multiple value function approximators could improve target estimates.
For partial observability, control-theoretic approaches to joint estimation and control may be promising.
Overall, \smpcritic offers a practical foundation for combining the online planning and constraint-handling of \ac{MPPI} with the adaptivity of \ac{RL}, enabling flexible control design.

\bibliographystyle{IEEEtran}
\bibliography{main.bib}

%\appendix

%\section{Experimental details}
%% stuff that doesn't make sense in the main body; maybe failed experiments too (baseline policy; uniform prior; value aligned model learning)
% other algorithmic expensions: utd; mppi adaptatin (cov/lambda) maybe for targets vs online mppi; double Q networks; terminations; training the model with different scales; episodic variants

\end{document}

%% file: acronyms.tex
\DeclareAcronym{AI}{short = AI,
	long = artificial intelligence}
\DeclareAcronym{ML}{short = ML,
	long = machine learning,
	short-indefinite = an}

\DeclareAcronym{SVD}{short = SVD, long = singular value decomposition}
\DeclareAcronym{CCA}{short = CCA,
	long = canonical correlation analysis}	
\DeclareAcronym{FA}{short = FA,
	long = factorial analysis}
\DeclareAcronym{GMM}{short = GMM,
	long = Gaussian mixture model}	
\DeclareAcronym{ICA}{short = ICA,
	long = independent component analysis}	
\DeclareAcronym{LARS}{short = LARS,
	long = least-angle regression}	
\DeclareAcronym{LASSO}{short = LASSO,
	long = least absolute shrinkage and selection operator}	
\DeclareAcronym{LR}{short = LR,
	long = logistic regression}
\DeclareAcronym{PCA}{short = PCA,
	long = principal component analysis}
\DeclareAcronym{PLS}{short = PLS,
	long = partial least squares}	
\DeclareAcronym{RBC}{short = RBC,
	long = reconstruction-based contribution}

\DeclareAcronym{ANFIS}{short = ANFIS,
	long = adaptive network fuzzy inference system}
\DeclareAcronym{ANN}{short = ANN,
	long = artificial neural network,
	short-indefinite = an}	
\DeclareAcronym{BN}{short = BN,
	long = Bayesian network}	
\DeclareAcronym{CNN}{short = CNN,
	long = convolutional neural network}	
\DeclareAcronym{DNNE}{short = DNNE,
	long = decorrelated neural network ensemble}		
\DeclareAcronym{DNN}{short = DNN,
	long = deep neural network}	
\DeclareAcronym{ELM}{short = ELM,
	long = extreme learning machine}
\DeclareAcronym{GAN}{short = GAN,
	long = generative adversarial network}
\DeclareAcronym{GPR}{short = GPR,
	long = Gaussian process regression}	
\DeclareAcronym{GRNN}{short = GRNN,
	long = general regression neural network}	
\DeclareAcronym{MLP}{short = MLP,
	long = multilayer perceptron,
	short-indefinite = an}
\DeclareAcronym{RBFNN}{short = RBFNN,
	long = radial basis function neural network,
	short-indefinite = an}
\DeclareAcronym{RNN}{short = RNN,
	long = recurrent neural network,
	short-indefinite = an}
\DeclareAcronym{RT}{short = RT,
	long = regression tree,
	short-indefinite = an}			
\DeclareAcronym{RVM}{short = RVM,
	long = relevance vector machine,
	short-indefinite = an}		
\DeclareAcronym{SFA}{short = SFA,
	long = slow feature analysis}	
\DeclareAcronym{SVM}{short = SVM,
	long = support vector machine}
\DeclareAcronym{TL}{short = TL,
	long = transfer learning}	
\DeclareAcronym{VAE}{short = VAE,
	long = variational autoencoder}	
\DeclareAcronym{WNN}{short = WNN,
	long = wavelet neural network}				

\DeclareAcronym{RL}{short = RL, long = reinforcement learning, short-indefinite = an}
\DeclareAcronym{A3C}{short = A3C,
	long = asynchronous advantage actor-critic}	
\DeclareAcronym{ADP}{short = ADP,
	long = approximate dynamic programming}
\DeclareAcronym{DP}{short = DP,
	long = dynamic programming}
\DeclareAcronym{IQC}{short = IQC, long = integral quadratic constraint}
\DeclareAcronym{BIBO}{short = BIBO, long = {bounded-input, bounded-output}}
\DeclareAcronym{SISO}{short = SISO, long = {single-input, single-output}}
\DeclareAcronym{MIMO}{short = MIMO, long = {multiple-input, multiple-output}}
\DeclareAcronym{DDPG}{short = DDPG,
	long = deep deterministic policy gradient}
\DeclareAcronym{DPG}{short = DPG,
	long = deterministic policy gradient}
\DeclareAcronym{DQN}{short = DQN,
	long = deep $Q$-network}
\DeclareAcronym{HJB}{short = HJB,
	long = {Hamilton-Jacobi-Bellman}}
\DeclareAcronym{MPC}{short = MPC, 
	long = model predictive control,
	long-plural-form = model predictive controllers,
	short-indefinite = an}		
\DeclareAcronym{MPPI}{short = MPPI, 
	long = model predictive path integral control,
	long-plural-form = model predictive path integral controllers,
	short-indefinite = an}	
\DeclareAcronym{PI2}{short = $\text{PI}^2$,
	long =policy improvement with path integrals}
\DeclareAcronym{PID}{short = PID, 
	long = proportional-integral-derivative}
\DeclareAcronym{PI}{short = PI, 
	long = proportional-integral}
\DeclareAcronym{PPO}{short = PPO,
	long = proximal policy optimization}
\DeclareAcronym{REINFORCE}{short = REINFORCE,
	long = {\emph{RE}ward Increment $=$ \emph{N}onnegative \emph{F}actor $\times$ \emph{O}ffset \emph{R}einforcement $\times$ \emph{C}haracteristic \emph{E}ligibility}}
\DeclareAcronym{RTO}{short = RTO,
	long = real-time optimization,
	short-indefinite = an}	
\DeclareAcronym{SAC}{short = SAC,
	long = soft actor-critic}	
\DeclareAcronym{TD3}{short = TD3,
	long = twin-delayed DDPG}	
\DeclareAcronym{HER}{short = HER,
	long = hindsight experience replay}	
\DeclareAcronym{DPC}{short = DPC,
	long = differentiable predictive control,
	long-plural-form = differentiable predictive controllers,
	short-indefinite = an}

\DeclareAcronym{GP}{short = GP,
	long = Gaussian process,
	long-plural-form = Gaussian processes}
\DeclareAcronym{MSE}{short = MSE,
	long = mean squared error}
\DeclareAcronym{RBF}{short = RBF,
	long = radial basis function,
	short-indefinite = an}	
\DeclareAcronym{SAE}{short = SAE,
	long = sparse autoencoder}	
\DeclareAcronym{DBN}{short = DBN,
	long = deep belief network}	
\DeclareAcronym{LSTM}{short = LSTM,
	long = long short-term memory}	
\DeclareAcronym{KL}{short = KL,
	long = Kullback-Leibler}
\DeclareAcronym{MDP}{short = MDP,
	long = Markov decision process,
	long-plural-form = Markov decision processes,
	short-indefinite = an}
\DeclareAcronym{LQR}{short = LQR, 
	long = linear quadratic regulator}
\DeclareAcronym{DARE}{short = DARE, 
	long = discrete algebraic Riccati equation}
\DeclareAcronym{LTI}{short = LTI, 
	long = linear time-invariant,
	short-indefinite = an}
\DeclareAcronym{GPS}{short = GPS,
	long = guided policy search}	
\DeclareAcronym{GRU}{short = GRU,
	long = gated recurrent unit}	
\DeclareAcronym{ESN}{short = ESN,
	long = echo state network}	
\DeclareAcronym{ENN}{short = ENN,
	long = Elman neural network}	
	
\DeclareAcronym{CSTR}{short = CSTR,
	long = continuous stirred tank reactor}